# Attentional Multilabel Learning over Graphs

## A Message Passing Approach

**Kien Do · Truyen Tran · Thin Nguyen · Svetha Venkatesh**

**Abstract** We address a largely open problem of multilabel classification over graphs. Unlike traditional vector input, a graph has rich variable-size substructures which are related to the labels in some ways. We believe that uncovering these relations might hold the key to classification performance and explainability. We introduce GAML (**G**raph **A**ttentional **M**ulti-**L**abel learning), a novel graph neural network that can handle this problem effectively. GAML regards labels as auxiliary nodes and models them in conjunction with the input graph. By applying message passing and attention mechanisms to both the label nodes and the input nodes iteratively, GAML can capture the relations between the labels and the input subgraphs at various resolution scales. Moreover, our model can take advantage of explicit label dependencies. It also scales linearly with the number of labels and graph size thanks to our proposed *hierarchical attention*. We evaluate GAML on an extensive set of experiments with both graph-structured inputs and classical unstructured inputs. The results show that GAML significantly outperforms other competing methods. Importantly, GAML enables intuitive visualizations for better understanding of the label-substructure relations and explanation of the model behaviors.

**Keywords** Multilabel learning · Graph classification · Graph neural networks · Message passing

This work is partially supported by the Telstra-Deakin Centre of Excellence in Big Data and Machine Learning.

Kien Do, Truyen Tran, Thin Nguyen and Svetha Venkatesh
Applied AI Institute, Deakin University, Australia
E-mail: {dkdo,truyen.tran,thin.nguyen,svetha.venkatesh}@deakin.ed.au

## 1 Introduction

Drug development costs billions of dollars over many years with multiple stages of refinement and trial [28]. For this reason, repurposing approved drugs is a critical alternative to the full development cycle, offering a huge saving of money, time, and lives. A canonical task in drug repurposing is to predict the drug effect on multiple related diseases. This can naturally be formulated as a multilabel learning problem. Different from standard domains like text or image, drugs are usually represented as variable size graphs of atoms linked by chemical bonds. The irregularity and complexity of rich graph structures make multilabel learning over molecular graph very challenging. At the same time, graph brings about new kinds of information not previously seen in unstructured data, as evidenced in recent surge of research in graph representation [19, 53]. Hence, a new proper treatment of the multilabeling over graphs is needed.

We hypothesize that the key for classification performance and explainability lies in uncovering the relations between labels and subgraphs. Towards this goal, we design a new graph neural network [35] called GAML (which stands for **G**raph **A**ttentional **M**ulti-**L**abel learning). GAML treats all label classes as nodes (termed label nodes) and merges them with other nodes (called input nodes) in the input graph to form a unified label-input graph. In the joint graph, relations between labels and substructures can be effectively captured through the interaction across the label nodes and the input nodes. Specifically, we leverage the message passing algorithm [31, 37, 14] to simultaneously update the local substructure at every input node and to propagate the substructure-contained messages from all the input nodes to the label nodes. By using attention [2, 51], each label node can extract the most related substructures to update its own state which will later be used to predict the presence of the corresponding class. Attention also enables insightful visualization which helps explain the prediction. To account for large number of classes and big input graph, we propose a new type of attention named *hierarchical attention*. Different from the standard approach that calculates the score matrix between every input and label node directly, our attention mechanism uses some intermediate *attentional factors* to save computation. In our model, implicit dependencies among the labels are captured via common attended substructures. However, when explicit dependencies among the labels are available (e.g, through expert knowledge), GAML can easily integrate them by adding links and exchange messages between related label nodes. Moreover, since the node update procedure runs iteratively, our model can learn the label-subgraph (or label-substructure) relations at various resolution scales.

The flexibility and scalability of GAML make it attractive to many real-world problems. In this paper, we focus on two major drug–multitarget prediction problems: predicting *drug–protein binding*, and *drug–cancer response*. In the first problem, a drug is tested against multiple target proteins; and in the second problem, a drug is tested against multiple cancer types. We also evaluate our method on classical vector input which can be seen as a special graph with a singleton node. In both cases, GAML proves to be superior against rival multilabel learning techniques. Finally, to get a clear picture of the learned label-substructure patterns, we generate visualizations using real drug molecules extracted from our datasets.

In summary, our contributions are:

– Proposing a novel neural graph neural network named GAML that addresses an open problem of multilabel classification over graphs. Our model can effectively capture the (multi-way) relations among the labels and the input subgraphs. It can also incorporate explicit label dependencies and is scalable to many labels and big graphs.
– Demonstrating the advantages of GAML through a comprehensive suit of experiments with quantitative evaluation and visualization.

## 2 Related Work

*Multilabel classification with label dependencies* Most work in multilabel learning focuses on capturing the implicit or explicit label dependencies. One strategy is applying Canonical Correlation Analysis (CCA) to map input and label into a common latent space. Then from this space, the model will reconstruct the target label. Extensions of this approach including both shallow [25,41] and deep [52] models. For graphical model-based approach, the work in [13] uses Conditional Random Fields to model the three way relation between every pair of labels $i$, $j$ and the input $\boldsymbol{x}$ using a feature function $\phi(y_i, y_j, \boldsymbol{x})$. Meanwhile, the work in [17] constructs a fully connected cyclic Bayesian Network over labels and perform structure learning on this network. The probability of a label $y_i$ conditioned on the input $\boldsymbol{x}$ and other labels $y_{\neg i}$ is modeled using a logistic regression network. Both methods are computationally expensive and require inexact inference for large number of labels.

To model the joint distribution of labels but still keep computation reasonable, some methods exploit chain rule factorization. The most notable one is Probabilistic Classifier Chain [9] which builds a separate binary classifier for each label with input to the model is the combination of the original input and the previously predicted labels. Other methods follow that idea but use recurrent neural networks [6,49] to learn the correlations better. However, the critical issues of these method are *ordering* and *poor inference* (since the the output label at one step depends on the value of the previous predicted labels not their distribution, which is very unstable). Although some tricks like beam search [49], or automatic order selection [6] are implemented to improve the results, they can only solve part of the problems.

Expert knowledge about label dependencies represented as trees [10] or graphs [3,5] has been exploited for multilabel/multiclass classification. In [5], the authors build a graph neural network over the predefined label graph. The input vector is copied for every label node and is concatenated to the label embedding vector to form an initial state for that label node. Their method, however, is limited to the vector input only whereas our model directly works on graph input with vector input is the special case.

*Multilabel classification with graph inputs* Although graph classification has attracted a significant interest in recent years [42], there has been a limited body of work on multilabel graph classification [23]. The line of work on image tagging considers multilabel learning over a grid of pixels [15,49,50]. However, the standard treatment using CNN usually focuses on attention over feature maps instead of exploiting the structural relations of objects in the original image. A recent work in visual question answering that pushes forward the idea of object graph is [43], but the QA setting is different from ours. A special case of our multilabel learning over graphs is multilabel learning over set [32] where input is a collection of nodes with no explicit links.

*Graph neural networks* By leveraging the representation power of deep neural networks such as CNN and RNN, a wide range of methods for learning over graphs [8,14,18,22,26,29,31,35] has been proposed recently. These methods can be grouped into more general categories such as Spectral Graph based [4,8,22], Message Passing based [14,31,37], Random Walk based [16,30], Neural Net based [26]. Among them, Message Passing Graph Neural Networks (MPGNNs) are very powerful since they can handle various kinds of graphs including attributed graphs whose edges and nodes both have types. MPGNNs have found many applications in bioinformatics such as drug activity classification [33], chemical properties prediction [14], protein interface prediction [11] and drug generation [20]. However, none of these methods properly handle multilabel classification problems in which modeling multi-way relations among labels and molecular subgraphs is the key factor.

## 3 Method

In this section, we present our main contribution–the GAML (Graph Attentional Multi-Label learning). First we provide background knowledge about graph neural networks, on which GAML is built.

### 3.1 Preliminaries: Message passing graph neural network

Consider an attributed graph $\mathcal{G} = (\mathcal{V}, \mathcal{E})$ where $\mathcal{V}$ is the set of nodes and $\mathcal{E}$ is the set of edges. Each node $i$ is associated with a feature vector $\boldsymbol{v}_i$ and each edge $(i, j)$ is associated with an attribute set $\boldsymbol{e}_{ij}$ (e.g., weight, type, features). In case node $i$ has type $t_i$, $\boldsymbol{v}_i$ can be an embedding vector of $t_i$. Let $\boldsymbol{x}_i$ be a state of node $i$ and $\mathcal{N}(i) = \{j \mid (i, j) \in \mathcal{E}\}$ denote the neighborhood of node $i$. In message passing graph neural networks [14,31,35], a node uses information from its neighbors to update its own state as follows:

$$\boldsymbol{x}_i^t \leftarrow f\left(\boldsymbol{x}_i^{t-1}, \left\{\left(\boldsymbol{x}_j^{t-1}, \boldsymbol{e}_{ij}\right)\right\}_{j\in\mathcal{N}(i)}\right) \tag{1}$$

where $t$ denotes update step; and $f(\cdot)$ is a non-linear function (e.g., a multilayer perceptron (MLP)). At $t = 0$, we set $\boldsymbol{x}_i^0 = \boldsymbol{v}_i$.

Eq. (1) is generic for most graph neural network models. In practice, it can be divided into two steps: *message aggregation* and *state update*. In the *message aggregation* step, we combine multiple messages sent to node $i$ into a single message vector $\boldsymbol{m}_i$:

$$\boldsymbol{m}_i^t = g^{\mathrm{a}}\left(\boldsymbol{x}_i^{t-1}, \left\{\left(\boldsymbol{x}_j^{t-1}, \boldsymbol{e}_{ij}\right)\right\}_{j\in\mathcal{N}(i)}\right) \tag{2}$$

where $g^{\mathrm{a}}(\cdot)$ can be an attention [2,51] or a pooling architecture. For example, the message aggregated using mean pooling has the following formula:

$$\boldsymbol{m}_i^t = \frac{1}{|\mathcal{N}(i)|}\sum_{j\in\mathcal{N}(i)} W_{e_{ij}}\boldsymbol{x}_j^{t-1} \tag{3}$$

for some parameter matrix $W_{e_{ij}}$. Despite of simplicity, Eq. (3) has shown to be able to encode graph structures in several message passing models [14,31,37].

During the *state update* step, the node state is updated as follows:

$$\boldsymbol{x}_i^t \leftarrow g^{\mathrm{u}}\left(\boldsymbol{x}_i^{t-1}, \boldsymbol{m}_i^t\right) \tag{4}$$

where $g^{\mathrm{u}}(\cdot)$ can be any type of deep neural networks such as MLP [22,18], RNN [35], GRU [26] or Highway Network [31]. In our model, we use Highway Network [40] for $g^{\mathrm{u}}(\cdot)$ as it has been shown to be effective for long range dependencies thanks to its skip-connection and gating mechanism. As a result, Eq. (4) now becomes:

$$\boldsymbol{x}_i^t \leftarrow \left(1 - \boldsymbol{\alpha}_i^t\right) \odot \boldsymbol{x}_i^{t-1} + \boldsymbol{\alpha}_i^t \odot \hat{\boldsymbol{x}}_i^t \tag{5}$$

where $\boldsymbol{\alpha}_i^t$ and $\hat{\boldsymbol{x}_i}^t$ are the gate vector and the non-linear candidate vector of node $i$ at time $t$, respectively; $\odot$ is the element-wise product. The formulas of $\boldsymbol{\alpha}^t$ and $\hat{\boldsymbol{x}}^t$ are provided below:

$$\begin{aligned}\boldsymbol{\alpha}_i^t &= \mathrm{sigmoid}\left(W_\alpha \boldsymbol{x}_i^{t-1} + U_\alpha \boldsymbol{m}_i^t + \boldsymbol{b}_\alpha\right)\\ \hat{\boldsymbol{x}}_i^t &= \mathrm{relu}\left(W_x \boldsymbol{x}_i^{t-1} + U_x \boldsymbol{m}_i^t + \boldsymbol{b}_x\right)\end{aligned}$$

where $W_\alpha, W_x U_\alpha, U_x$, and $\boldsymbol{b}_\alpha, \boldsymbol{b}_x$ are parameters which can be different or shared among layers. During experiments, we observed that models with parameter sharing run faster but still provide comparable results. Hence, we applied this sharing scheme to our model. We abstract Eqs. (4,5) into:

$$\boldsymbol{x}_i^t \leftarrow \mathrm{Highway}\left(\boldsymbol{x}_i^{t-1}, \boldsymbol{m}_i^t\right) \tag{6}$$

After $T$ steps of message passing, $\boldsymbol{x}_i^T$ would capture the graph substructure centered at node $i$ with radius $T$. The graph summary vector $\boldsymbol{x}_\mathcal{G}$ is the combination of the state vector of all nodes in the graph at step $T$. In the simplest form, $\boldsymbol{x}_\mathcal{G}$ is the average of $\left\{\boldsymbol{x}_i^T\right\}_{i\in\mathcal{V}}$, as follows:

$$\boldsymbol{x}_\mathcal{G} = \frac{1}{|\mathcal{V}|}\sum_{i\in\mathcal{V}} \boldsymbol{x}_i^T$$

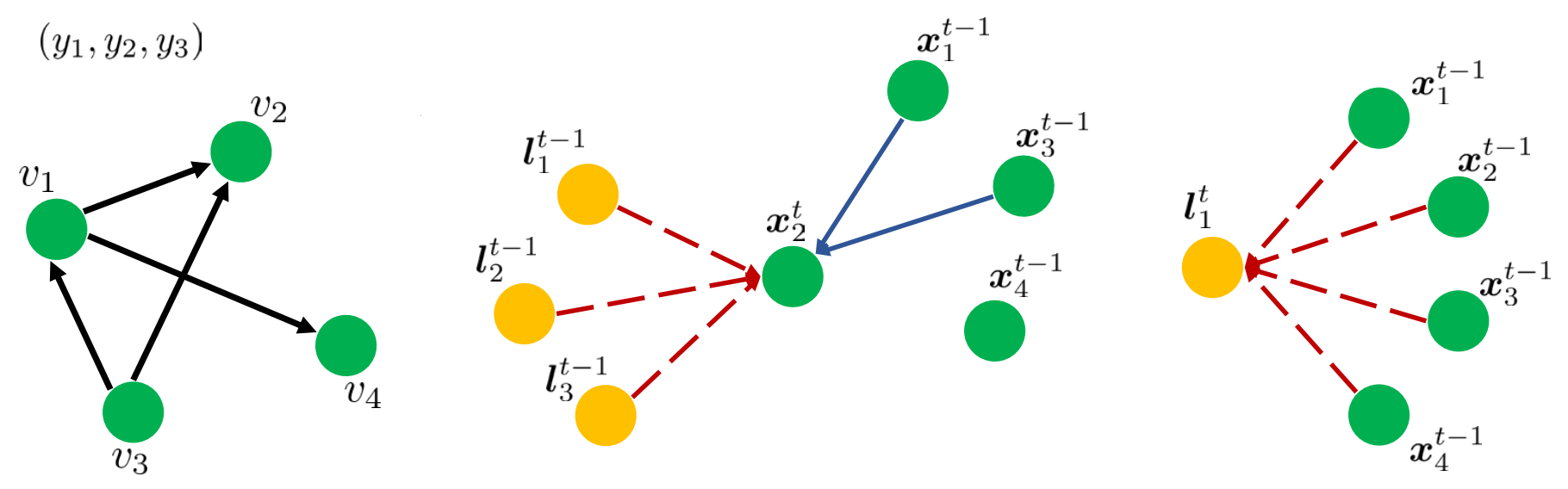

(a) A input graph with 4 nodes and 3 labels

(b) Input node update

(c) Label node update

Fig. 1: Message passing in the joint graph of input nodes and label nodes. In (b) and (c), dash red link indicates message passing with attention while blue solid link indicates message passing with mean pooling

### 3.2 Multilabel learning over graphs

Multilabel graph classification associates a graph $\mathcal{G}$ with a label vector $\boldsymbol{y} = (y_1, ..., y_C) \in \{0,1\}^C$ where $y_c = 1$ indicates that class $c$ is present given $\mathcal{G}$. We argue that in order to perform well on this task, two correlation structures must be captured: those *within the label set* and those *between the label set and input subgraphs.*

Let us start with representing labels in the same vector space. When prior label descriptors exist, the space is simply the descriptor vector space. Otherwise, labels are embedded using a *class embedding matrix* $M \in \mathbb{R}^{C \times d_l}$ as follows:

$$\boldsymbol{l}_c^0 = M \ \mathbf{1}_c$$

where $\mathbf{1}_c \in \mathbb{R}^C$ is an one-hot vector with the $c$-th element equal to 1.

For an input graph $\mathcal{G}$, we then consider all the $C$ classes as auxiliary nodes (called *label nodes*) alongside $|\mathcal{V}|$ existing nodes of the input graph $\mathcal{G}$. Each label node $c$ has the initial state $\boldsymbol{l}_c^0$ and connects to all input nodes. Similarly, each input node $i$ also connects to all label nodes. It results in a joint graph of $C + |\mathcal{V}|$ nodes, which naturally lend itself to the message passing scheme in the graph neural network presented in Section 3.1. The idea is that by iteratively updating the states of input and label nodes using message passing, complex label-label and label-substructure dependencies emerge. See Fig. 1 for an illustration.

### 3.2.1 Input node update

Since an input node $i$ connects to its neighbor nodes $j \in \mathcal{N}(i)$ and all the label nodes $c \in \overline{1,C}$ , the message passing update of the input node $i$ at step $t$ is formulated as follows:

$$\boldsymbol{x}_i^t = f\left(\boldsymbol{x}_i^{t-1}, \left\{\left(\boldsymbol{x}_j^{t-1}, \boldsymbol{e}_{ij}\right)\right\}_{j\in\mathcal{N}(i)}, \left\{\boldsymbol{l}_c^{t-1}\right\}_{c\in\overline{1,C}}\right) \tag{7}$$

Note that Eq. (7) is derived from Eq. (1) with the introduction of new arguments $\left\{\boldsymbol{l}_c^{t-1}\right\}_{c\in\overline{1,C}}$.

There are two types of message sent to the input node $i$. One contains structure information from neighbor input nodes and the other contains label-related information from label nodes. Because these messages have different meanings, they should be aggregated into separate message vectors. In case of neighbor input nodes, we use mean pooling to combine them as similar to Eq. (3):

$$\boldsymbol{\mu}_i^t = \frac{1}{|\mathcal{N}(i)|}\sum_{j\in\mathcal{N}(i)} W_{e_{ij}}\boldsymbol{x}_j^{t-1}$$

However, mean pooling may not be ideal to aggregate labels since it equalizes the importance of each class towards the input node $i$. To overcome this issue, we use the attention mechanism [2,51] to compute a weighted sum of all the label nodes as follows:

$$\boldsymbol{m}_i^t = \sum_{c=1}^{C} a_{ic}^t \boldsymbol{l}_c^{t-1} \tag{8}$$

where $a_{ic}^t > 0$, $\sum_{c=1}^{C} a_{ic}^t = 1$ is the attention coefficient from the input node $i$ to a label node $c$ at time $t$, computed as:

$$s_{ic}^t = \boldsymbol{u}_s^\intercal \tanh\left(W_s\boldsymbol{x}_i^{t-1} + U_s\boldsymbol{l}_c^{t-1} + \boldsymbol{b}_s\right) \tag{9}$$

$$a_{ic}^t = \frac{\exp\left(s_{ic}^t\right)}{\sum_{c'=1}^{C}\exp\left(s_{ic'}^t\right)} \tag{10}$$

The set of all unnormalized attention scores $s_{ic}^t$ in Eq. (9) forms a matrix $S^t \in \mathbb{R}^{|\mathcal{V}|\times C}$, which we will reuse later.

For generality, Eqs. (8–10) are written in a more compact form:

$$\boldsymbol{m}_i^t = \text{Attention}\left(\boldsymbol{x}_i^{t-1}, \left\{\boldsymbol{l}_c^{t-1}\right\}_{c\in\overline{1,C}}\right) \tag{11}$$

We call the attention in Eq. (11) *input-to-label* attention.

In the state update phase, the new state $\boldsymbol{x}_i^t$ of the input node $i$ is computed as:

$$\boldsymbol{x}_i^t = \text{Highway}\left(\boldsymbol{x}_i^{t-1}, \left[\boldsymbol{\mu}_i^t, \boldsymbol{m}_i^t\right]\right)$$

where $[\cdot]$ denotes vector concatenation and Highway() is defined in Eq. (6).

*3.2.2 Label node update*

By connecting to every input node, a label node $c$ can receive information about various substructures in the graph $\mathcal{G}$ through multiple steps of message passing. Among these substructures, only a few are related to the class $c$. Therefore, we use the attention mechanism to extract the most useful substructures for predicting class $c$ and store them in the message vector as follows:

$$\boldsymbol{m}_c^t = \text{Attention}\left(\boldsymbol{l}_c^{t-1}, \left\{\boldsymbol{x}_i^{t-1}\right\}_{i\in\overline{1,|\mathcal{V}|}}\right) \tag{12}$$

where Attention(.) is similar to the function defined in Eq. (11) with the role of input nodes and label nodes swapped. We denote this function *label-to-input* attention. The unnormalized score matrix $S^t$ from Eq. (9) is reused here to save computation and improve consistency. However, the attention coefficients are be normalized over rows instead of columns of $S^t$, i.e.,

$$a_{ci}^t = \frac{\exp\left(s_{ic}^t\right)}{\sum_{i=1}^{|\mathcal{V}|}\exp\left(s_{ic}^t\right)}$$

Finally, we compute the new state of the label node $c$ using a different Highway Network as:

$$\boldsymbol{l}_c^t = \text{Highway}\left(\boldsymbol{l}_c^{t-1}, \boldsymbol{m}_c^t\right)$$

*3.2.3 A priori label dependencies*

When explicit label dependencies are available, a label graph can be formed in the same way as the input graph. Messages between label nodes is aggregated using mean-pooling as in Eq. (3):

$$\boldsymbol{\mu}_c^t = \frac{1}{|\mathcal{N}(c)|}\sum_{f\in\mathcal{N}(c)} W_{e_{cf}}\boldsymbol{l}_f^{t-1}$$

The state of the label node $c$ is updated as:

$$\boldsymbol{l}_c^t = \text{Highway}\left(\boldsymbol{l}_c^{t-1}, \left[\boldsymbol{m}_c^t, \boldsymbol{\mu}_c^t\right]\right)$$

*3.2.4 Vector input as a special case*

In many traditional multilabel classification problems, the input is represented as vector instead of graph. This can be seen as a special case of our model where the input graph $\mathcal{G}$ collapses into a single node $\boldsymbol{x}$. With this observation, the state update of the input node at step $t$ is:

$$\begin{aligned}\boldsymbol{m}^t &= \text{Attention}\left(\boldsymbol{x}^{t-1}, \left\{\boldsymbol{l}_c^{t-1}\right\}_{c\in\overline{1,C}}\right)\\ \boldsymbol{x}^t &= \text{Highway}\left(\boldsymbol{x}^{t-1}, \boldsymbol{m}^t\right)\end{aligned}$$

The state of a label node $c$ is updated as:

$$\boldsymbol{l}_c^t = \text{Highway}\left(\boldsymbol{l}^{t-1}, \boldsymbol{x}^{t-1}\right)$$

*3.2.5 Learning*

After $T$ steps of message passing, we pass each class-specific final state vector $\boldsymbol{l}_c^T$ to a multi-layer perceptron (MLP) with sigmoid activation on top to predict the present of class $c$:

$$o_c = \text{MLP}\left(\boldsymbol{l}_c^T\right)$$

Here the value of $o_c$ is in $(0, 1)$. The MLPs for all classes share the same parameters. For learning, we use a binary cross-entropy loss function which is defined as:

$$\mathcal{L} = \mathbb{E}_{\text{train}}\left(\sum_{c=1}^{C} y_c \log o_c + (1 - y_c)\log(1 - o_c)\right)$$

where $\mathbb{E}_{\text{train}}$ denotes the mean over all training data.

## 3.3 Scale to big graphs and many labels

When the number of nodes in the input graph ($|\mathcal{V}|$) and the number of classes ($C$) are large, it becomes expensive to calculated the unnormalized score matrix $S^t \in \mathbb{R}^{|\mathcal{V}| \times C}$ in Eq. (9) for all steps $t = 1, 2, ..., T$. To handle this problem, we propose a new attention technique called *hierarchical attention*. At each layer, we define $K$ ($K \ll \min\{|\mathcal{V}|,\ C\}$) intermediate *attentional factors* between input nodes and label nodes. The input-label attentions are broken down into two steps as follows:

- For *label-to-input* attention, we do *factor-to-input* attention then *label-to-factor* attention.
- For *input-to-label* attention, we do *factor-to-label* attention then *input-to-factor* attention.

*Label-to-input message aggregation.* More concretely, the *label-to-input* message aggregation in Eq. (8) is replaced by:

$$\boldsymbol{m}_i^t = \sum_{k=1}^{K} a_{ik}^t \boldsymbol{\lambda}_k^{t-1}; \quad \text{for} \quad \boldsymbol{\lambda}_k^{t-1} = \sum_{c=1}^{C} b_{ck}^t \boldsymbol{l}_c^{t-1}$$

where $\boldsymbol{\lambda}_k^{t-1}$ is the k-th intermediate factor ($k \in \overline{1, K}$) that aggregates all label nodes; $\boldsymbol{m}_i^t$ is the message to the input node $i$; $a_{ik}^t$ is factor-to-input attention probability (i.e., $\sum_{k=1}^{K} a_{ik}^{t} = 1$); and $b_{ck}^{t}$ is label-to-factor attention probability (i.e., $\sum_{c=1}^{C} b_{ck}^{t} = 1$).

To compute $a_{ik}^{t}$ and $b_{ck}^{t}$ we define two score matrices $S_1^t \in \mathbb{R}^{|\mathcal{V}|\times K}$ and $S_2^t \in \mathbb{R}^{C\times K}$ as follows:

$$s_{1;ik}^{t} = \boldsymbol{u}_1^{\intercal} \tanh(W_1 \boldsymbol{x}_i^{t-1} + \boldsymbol{z}_k^{t-1}) \tag{13}$$

$$\text{and} \quad s_{2;ck}^{t} = \boldsymbol{u}_2^{\intercal} \tanh(W_2 \boldsymbol{l}_c^{t-1} + \boldsymbol{z}_k^{t-1}) \tag{14}$$

where $\boldsymbol{u}_1, \boldsymbol{u}_2 \in \mathbb{R}^{d_z}$, $W_1 \in \mathbb{R}^{d_x \times d_z}$, $W_2 \in \mathbb{R}^{d_l \times d_z}$ and $\boldsymbol{z}_k^t \in \mathbb{R}^{d_z}$, $(k = \overline{1,K})$ are parameters. Then factor-to-input attention probability and label-to-factor attention probability are computed as:

$$a_{ik}^{t} = \frac{\exp(s_{1;ik'}^{t})}{\sum_{k'=1}^{K} \exp(s_{1;ik'}^{t})}; \quad b_{ck}^{t} = \frac{\exp(s_{2;c'k}^{t})}{\sum_{c'=1}^{C} \exp(s_{2;c'k}^{t})}$$

*Input-to-label message aggregation.* Likewise the two-step *input-to-label* message aggregation is computed as:

$$\boldsymbol{m}_c^t = \sum_{k=1}^{K} \alpha_{ck}^{t} \boldsymbol{\chi}_k^{t-1}; \quad \text{for} \quad \boldsymbol{\chi}_k^{t-1} = \sum_{i=1}^{|\mathcal{V}|} \beta_{ik}^{t} \boldsymbol{x}_i^{t-1}$$

where $\boldsymbol{\chi}_k^{t-1}$ is the k-th intermediate factor $(k \in \overline{1,K})$ that aggregates all input nodes; $\boldsymbol{m}_c^t$ is the message to the label node $c$; $\alpha_{ck}^{t}$ is factor-to-label attention probability (i.e., $\sum_k \alpha_{ck}^{t} = 1$); and $\beta_{ik}^{t}$ is input-to-factor attention probability (i.e., $\sum_i \beta_{ik}^{t} = 1$). The attention probabilities are respectively computed as:

$$\alpha_{ik}^{t} = \frac{\exp(s_{1;ik}^{t})}{\sum_{i'=1}^{|\mathcal{V}|} \exp(s_{1;i'k}^{t})}; \quad \beta_{ck}^{t} = \frac{\exp(s_{2;ck}^{t})}{\sum_{k'=1}^{K} \exp(s_{2;ck'}^{t})}$$

where the scores $s_{1;ik}^{t}$ and $s_{2;ck}^{t}$ are computed using Eqs. (13,14).

It is clear that with this decomposition strategy, the number of computation steps reduces from $\mathcal{O}(|\mathcal{V}|C)$ to $\mathcal{O}((|\mathcal{V}| + C)K)$ for $K \ll \min\{|\mathcal{V}|,\ C\}$.

### 3.4 Detecting higher-order correlation

*Higher-order label correlation* The iterative message passing scheme spreads information to distant nodes. Two labels can indirectly interact with each other after two step of updates: a label sends messages to input nodes which then redistribute the information back to other labels. This brings about higher-order label correlation.

*Multi-resolution substructure-label correlation* Likewise, after $t$ steps, an input node accumulates information from other nodes of $t$ degrees of separation. In other words, the node state represents a latent subgraph of radius $t$. Thus, input-to-label attention detects substructures of the input graph with varying resolution.

## 4 Experiments

We present empirical results on two comprehensive sets of experiments: one on graph-structured input (Section 4.1) and the other on traditional unstructured input (Section 4.2).

### 4.1 Multilabel classification with graph-structured input

Our experiments focus on biochemical databases of potential drugs. A drug is a moderate-sized molecule with desirable bioactivities treated as labels. In the molecular graph of a drug, nodes represent atoms and edges represent bond types.

#### *4.1.1 Datasets*

We use two real-world biochemical datasets:

- *9cancers*: For this dataset, the goal is to predict drug activity against nine types of cancer (see Table. 1). The activity is binary indicating whether there is a response, i.e., the drug reduces or prevents tumor growth. We first download nine separate datasets for each cancer type from PubChem[1]. Then, we search for drug molecules that appear in all datasets, which results in about 22 thousand molecules in total. Among them, there are 3,356 molecules active for at least one type of cancer. We select all the active molecules and 10,000 fully inactive molecules to create the final dataset for experiment.
- *50proteins*: This dataset is about drug-protein binding prediction. Again, drugs are treated as input graphs while proteins are labels. We obtain the raw version from BindingDB[2]. In this dataset, the number of unique proteins (also called targets) is 595 and the number of unique drugs (or ligands) is 55,781. We select top 50 proteins that are bound by most ligands to construct our experimental dataset. There are 36,349 ligands in total with the average number of proteins to which one ligand binds is 1.35.

We divide each dataset into train/valid/test sets with the proportions of 0.6/0.2/0.2, respectively. The detailed statistics are shown in Table. 2 and the number of label occurrences is shown in Fig. 2. The labels in *50proteins* are sparse as each ligand links to at most 10 proteins (but the majority of ligands bind to only 1 or 2 proteins). Meanwhile, the labels in *9cancers* are denser with nearly a thousand of drugs positive to all cancers.

[1] https://pubchem.ncbi.nlm.nih.gov/
[2] http://www.bindingdb.org/bind/index.jsp

| Assay ID | Cancer Type | %Positive |
|---|---|---|
| 1 | Lung | 12.28 |
| 33 | Melanoma | 9.97 |
| 41 | Prostate | 11.77 |
| 47 | Central Nervous System | 12.22 |
| 81 | Colon | 14.50 |
| 83 | Breast | 16.22 |
| 109 | Ovarian | 12.76 |
| 123 | Leukemia | 18.91 |
| 145 | Renal | 12.03 |

Table 1: Assay ID and name of nine cancers in *9cancers* dataset extracted from PubChem. %Positive denotes the average percentage of positive examples for each cancer type over the total number of 13,356 molecules.

| Dataset | #labels | avg. #nodes | avg. #edges | #node types | #edge types |
|---|---|---|---|---|---|
| *9cancers* | 9 | 27.68 | 29.95 | 43 | 4 |
| *50proteins* | 50 | 25.31 | 27.49 | 14 | 4 |

Table 2: Statistics of all multilabel datasets with graph-structured inputs.

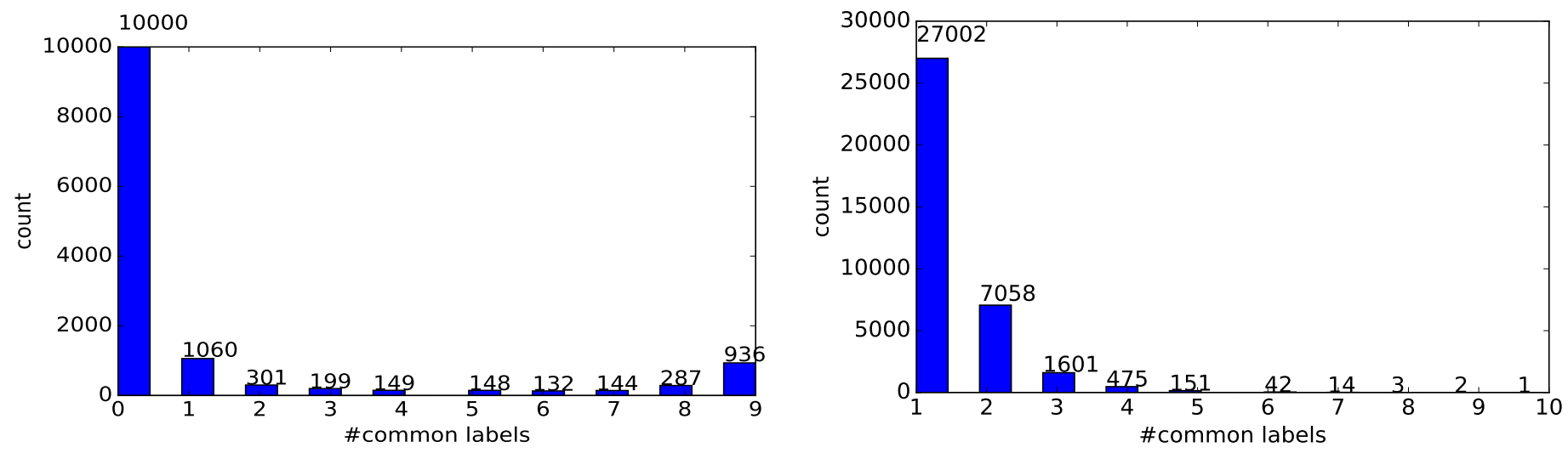

(a) 9cancers (b) 50proteins

Fig. 2: Histogram of the number of common labels that each instance associates to in *9cancers* and *50proteins*

### *4.1.2 Baselines*

For comparison, we employ the following data representations and associated multilabel classifiers:

*Molecular fingerprint.* The first set of baselines works on *molecular fingerprints*. A molecular fingerprint is a binary vector whose each element is associated with a particular type of substructures in the molecular graph. We use the well-known Morgan algorithm from RDKit[3] to generate multiple fingerprints with an increasing radius from 1 to 5 to account for fine-grained levels of substructures. Then, these fingerprints are concatenated to form a final feature vector. For each radius, we set the length of the fingerprint hash vector to 100. This results in the final feature vector of size 500. We evaluate two models running on top of this vector representation:

– The first model is a SVM with RBF kernels set as a base classifier for Binary Relevance algorithm [45]. We denote this model as fp+SVM.
– The second model is a Highway Network (HWN) [40] followed by a fully connected neural network with sigmoid activation function. All highway layers share parameters. We denote the combination of fingerprint and HWN as fp+HWN. In this model, the dependencies among classes are implicitly captured through the intermediate hidden layers.

*String representation.* SMILES is one of the most popular string representation of molecules which encapsulates the graph structure into its grammar. We consider SMILES as a sequence of characters and model it using a GRU [7]. When reaching the end of the sequence, the last state of the GRU is fed to a 2-layer MLP that outputs prediction for all labels. This SMILES+GRU combination has been recently proven to be highly effective in drug evaluation and design [38].

*Graph representation.* The last set of baselines handle graph-structured input directly. We select two representative models: Weisfeiler-Lehman Graph Kernels (WLs) [39] for graph kernel based methods and Column Networks (CLNs) [31] for graph neural network based methods.

– For WLs, we precompute the kernel matrix for both training and testing data using WL algorithm. The height of the rooted tree is chosen to be 3. For *9cancers*, it results in about 49 thousand different tree structures for all nodes in the graph dataset. Meanwhile, the total number of graphs is only about 13 thousands. Therefore, increasing height more than 3 will add very little information about graph similarity as the proportion of matching substructures approach zero. The kernel matrix for training graphs is used as input to a SVM wrapped by Binary Relevance (WL+BR).
– For CLNs, we use the same model as in [31] with a mean pooling layer on top message passing layers to compute the graph representation vector. This vector is then fed to a 2-layer MLP to predict all labels. Different from our GAML, a CLN only captures the relations between the labels and the subgraphs at the topmost layer rather than at every layer.

The hyper-parameters of fp+HWN and SMILES+GRU are obtained through validation. Meanwhile, the hyper-parameters of CLN are set similar to the optimal hyper-parameters of our model (see below).

[3] http://www.rdkit.org/

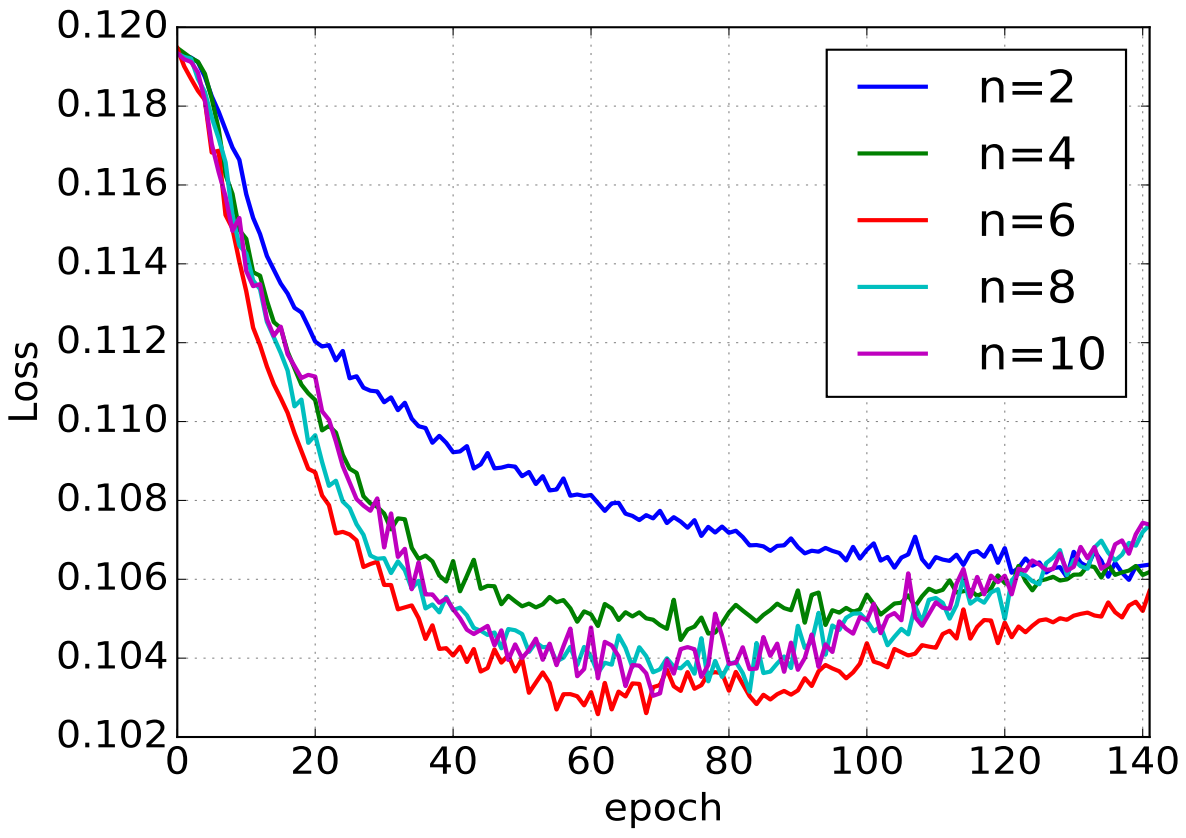

Fig. 3: Learning curves on *50proteins* with different number of message passing layers $n \in \{2, 4, 6, 8, 10\}$. Best viewed in color.

#### *4.1.3 Model setting*

In our model, the sizes of the node and edge embedding are both set to 50. We perform grid search for other hyper-parameters with the label embedding size in $\{10, 30, 50, 70, 100\}$, the number of factors in $\{1, 5, 10, 15, 20\}$, and the number of message passing layers in $\{2, 4, 6, 8, 10\}$. Dropout is set for every graph input node with the rate of 0.3. We do not use dropout for label nodes as it results in low F1 score although it makes the model less overfitting. In addition, we set the batch size to 60 and 100 for *9cancers* and *50proteins*, respectively. We use Adam optimizer [21] with an initial learning rate of 0.001. During training, the learning rate will be reduced by half if the validation loss does not improve after 20 consecutive epochs. We train our model for a maximum of 300 epochs and may stop early after decaying the learning rate 4 times.

#### *4.1.4 Evaluation metrics*

We use popular metrics for multilabel classification which are micro, macro (sometimes called per label) F1 and micro, macro AUC. While micro F1 favors labels with many examples due to its global averaging, macro F1 treats all labels equally regardless of their sample size, hence, is a good indication of the model performance on small labels.

#### *4.1.5 Parameter sensitivity*

To have a deep understanding of how GAML works for graph structured input, we investigate the contribution of different hyper-parameters including: the number of message passing layers (Fig. 3 and Fig. 4), the number of attention factors (Fig. 6), and the type of attention (Fig. 5). We report results for *50proteins*, but similar results are also observed for *9cancers*.

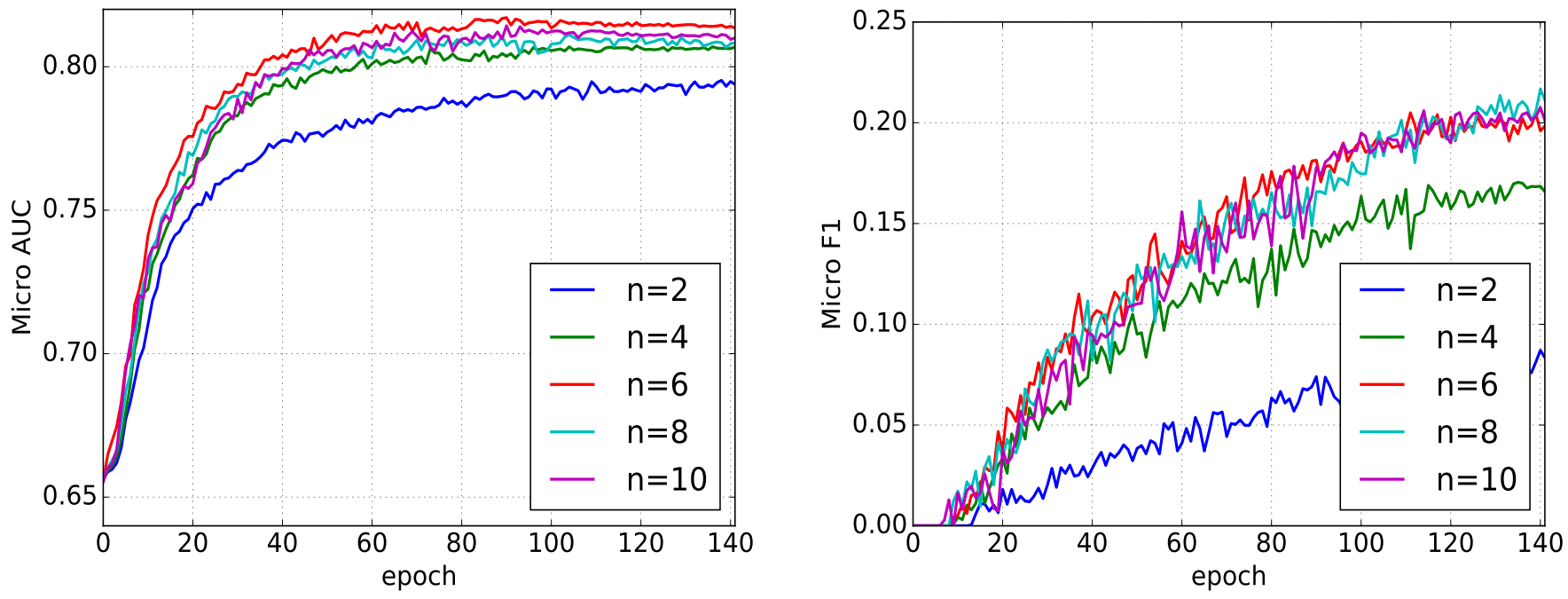

(a) Micro AUC (b) Micro F1

Fig. 4: Micro AUC (a) and Micro F1 (b) on 50proteins with different number of message passing layers $n \in \{2, 4, 6, 8, 10\}$. Best viewed in color.

From Fig. 4, it is seen that when the number of layers $n$ is small, e.g. $n = 2$, the model performs sub-optimally. Increasing the number of layers usually improves the results. We hypothesize that at higher level, input nodes receive a wider range of structural information through message passing. However, when $n \geq 6$, the improvement rate becomes steady and the model is more likely to overfit (see Fig. 4c). We believe there are two reasons for this situation: (i) the structure information from distant nodes is much less important than that from close neighbors; and (ii) the structure information at every node becomes more global and indistinguishable, causing difficulty for the model to detect meaningful substructures during prediction.

Another factor that affects the model performance is the type of attention. Generally, using attention provides better micro F1 score than not using it. However, the input-to-label attention seems to be redundant and causes misleading to the model. We observed that when the input-to-label attention is available, the model often has higher loss and lower micro AUC (see Fig. 5a and c). Meanwhile, the label-to-input attention is important as it helps the label nodes focus on particular substructures of the input graph to give accurate prediction. One interesting thing to note here is that the improvement of micro F1 by using attention mainly comes from micro Recall (as can be seen from Fig. 5b, d, e) and since the denominator in the micro Recall formula is constant (which is equal to the number of positive examples in the dataset), the number of true positives actually increases.

GAML performs worst in term of both micro AUC and micro F1 when the number of attention factors $k$ is 1 which is equivalent to collapsing all the neighbor nodes into one aggregating vector. For other values from 10 to 20, the results are quite comparable, which suggests that a small value of $k$ is usually sufficient.

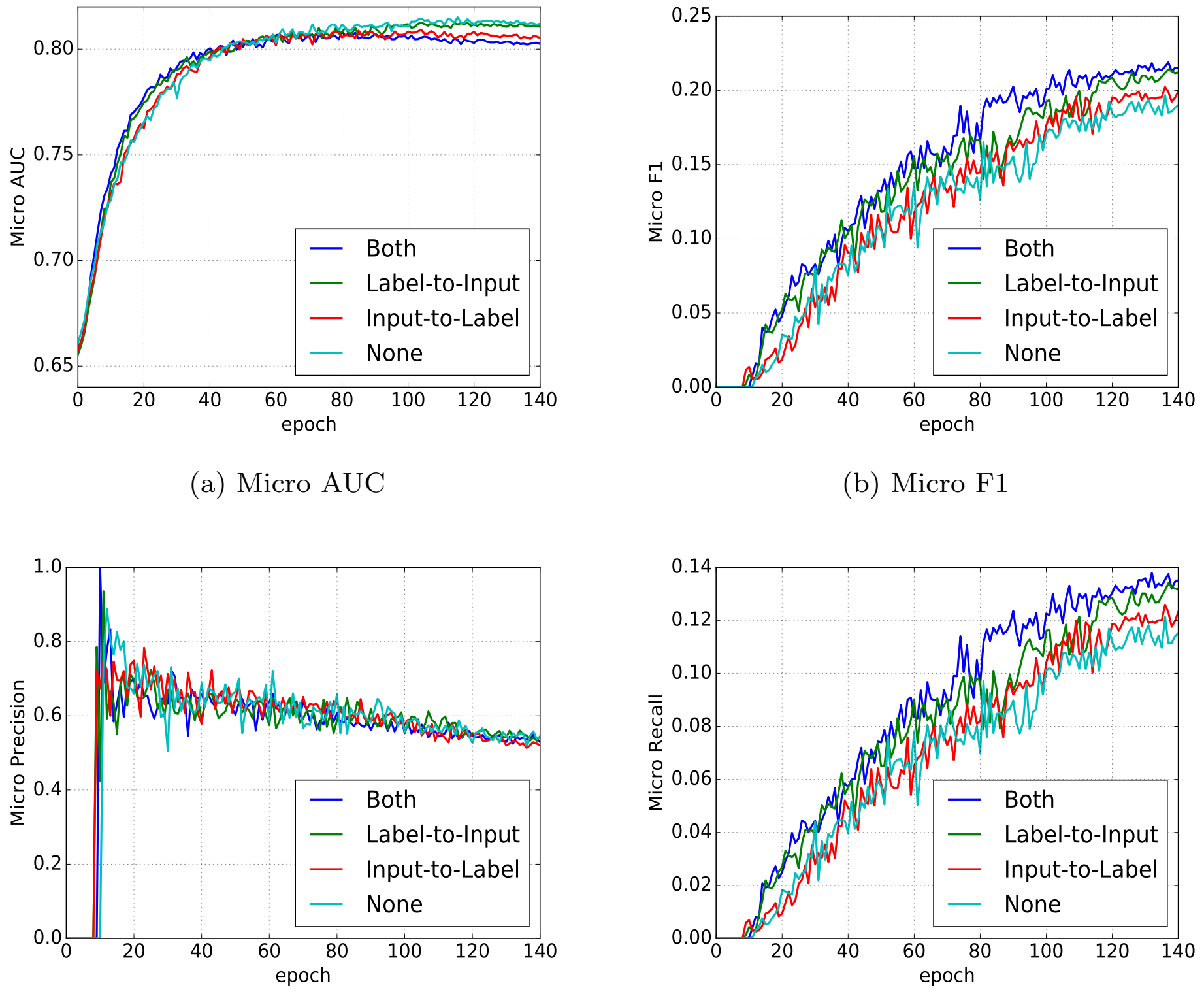

(a) Micro AUC (b) Micro F1

(c) Micro Precision (d) Micro Recall

Fig. 5: Results on *50proteins* with different type of attentions. *Label-to-Input* refers to unidirectional attention from label to input nodes; *Input-to-Label* refers to attention in the reverse direction; *Both* refers to bidirectional attention. Best viewed in color.

### *4.1.6 Performance results*

Table 3 shows the classification results for graph structured input. GAML consistently beats all baselines on all evaluation metrics. In particular, our model achieves about 2%-3% higher F1 and about 0.25%-1% higher AUC than the second best method (CLN) on both datasets. We believe this improvement comes from the fact that our model can associate labels with useful multi-resolution substructures of the input graph through attention mechanism while CLN does not have this capability. Furthermore, it is also clear that the models learning directly on graphs such as WL+BR or CLN usually provide better results than those learning on strings or vectors. For example, CLN achieves roughly 2% improvement in term of micro and macro F1 compared to its vector counterpart fp+HWN. Whereas, WL+BR produces about 2-4% higher macro and micro AUC than fp+SVM+BR.

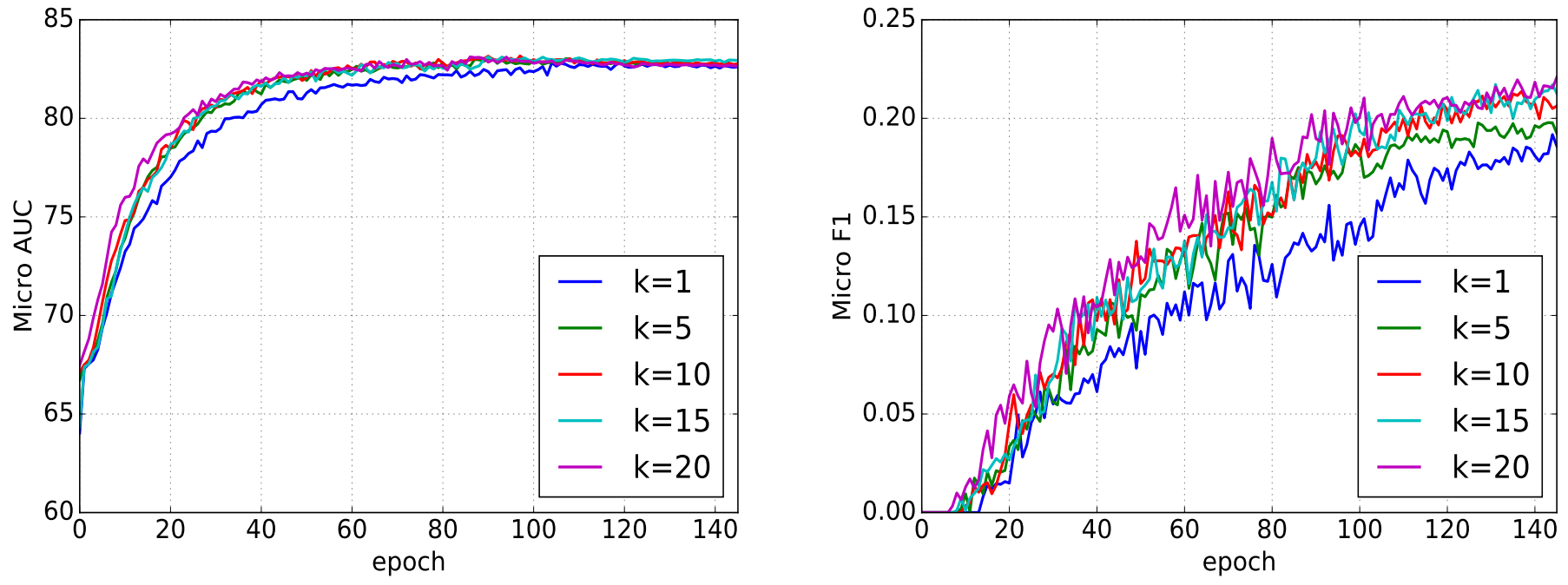

(a) Micro AUC (b) Micro F1

Fig. 6: Micro AUC (a) and Micro F1 (b) on 50proteins with different number of factors $k \in \{1, 5, 10, 15, 20\}$. Best viewed in color.

| Dataset | Metrics | SVM | HWN | GRU | WL+BR | CLN | GAML |
|---|---|---|---|---|---|---|---|
| *9cancers* | m-AUC | 81.94 | 85.95 | 83.29 | 86.06 | 88.35 | **88.78** |
| | M-AUC | 81.37 | 85.85 | 82.74 | 85.74 | 88.23 | **88.50** |
| | m-F1 | 50.63 | 57.44 | 55.97 | 54.55 | 59.48 | **62.03** |
| | M-F1 | 50.71 | 57.29 | 55.99 | 54.54 | 59.50 | **62.14** |
| *50proteins* | m-AUC | 79.85 | 77.46 | 79.11 | 81.62 | 82.08 | **82.82** |
| | M-AUC | 74.77 | 73.78 | 75.25 | 77.60 | 78.36 | **79.35** |
| | m-F1 | 17.21 | 16.37 | 16.08 | 17.04 | 18.37 | **20.47** |
| | M-F1 | 18.40 | 15.87 | 14.96 | 18.66 | 17.72 | **19.83** |

Table 3: The performance in the multi-label classification with graph-structured input (m-X: micro average of X; M-X: macro average). SVM and HWN work on fingerprint representation; GRU works on string representation of molecule known as SMILES; WL+BR and CLN work directly on graph representation. Bold indicates better values.

### *4.1.7 External knowledge of label dependencies*

*A priori* label dependencies are known to improve model performance as they bring structural constraints to the output space [24, 44]. We consider the setting where label dependencies form a graph. The multilabeling becomes node classification in the label graph *conditioned on the input graph*. We investigate the case of *50proteins* where the labels are sparse. We compute the protein-protein interaction (PPI) scores by using Human Integrated Protein-Protein Interaction rEference (HIPPIE) [1]. HIPPIE provides a normalized scoring scheme that integrates multiple PPI sources [36], hence, is reliable. The PPI scores have already been normalized in the range of [0, 1]. We add an edge between two proteins if their interaction score is larger than a predefined threshold (which set to 0.5 in our experiment). Since the interaction scores are asymmetric, the edges are directed. Table 4 reports results of our model when external label dependencies are introduced. The results are improved on F1 measures but not on the AUC scores suggesting that the external label constraints may help balance recall and precision when labels are sparse.

| Model | *50proteins* | | | |
|---|---|---|---|---|
| | m-AUC | M-AUC | m-F1 | M-F1 |
| GAML | **82.82** | **79.35** | 20.47 | 19.83 |
| GAML + PPI | 82.61 | 79.29 | **21.15** | **20.28** |

Table 4: Results on incorporating external knowledge of label dependencies. Bold indicates better values.

#### *4.1.8 Attention visualization*

In Fig. 7, we show the label-to-input attention scores at different message passing layers when our model runs on *9cancers* to see how our model matches labels to substructures of the input graph. At the first layer, the label nodes often attend to many input nodes. The reason is that input nodes at this level only contain information about their types. In addition, the attended input nodes are usually special atoms like Oxygen (8) or Nitrogen (7) instead of the common Carbon (6). However, the attention becomes more focused when going up to higher layer since the structure information at each input node has been updated via message passing. Sometimes, new substructures emerge and the model may switch its attention to these substructures if it finds them to be more appropriate.

From the label-to-input attention matrices in Fig. 7, we can map back to the molecule graph to detect meaningful substructures toward labels. In Fig. 8, we can observe the shift in the model attention with respect to the evolution of structures across layers. In particular, at layer 2, the model focuses most on the O-N substructure. However, at layer 3, the model changes its attention to N[6], N[5] and C[11] instead of Os. The reason is that the model becomes more interested in the appearance of two adjacent Ns in an aromatic group, which cannot be captured within two hops by starting at O. Therefore, an *attention shift* is performed by the model.

Note that although the attention shift looks disruptive in Fig. 7 as the model is highly attentive (due to well training), it is actually smooth under graphical view in Fig. 8 since the substructures rooted at N[6], N[5] and C[11] all contains the substructure O-N from the previous layer. This suggests a human-like concept transferring mechanism through attention where the old concept is not totally discarded but still exists as part of the new concept with less focusing from the brain. From layer 3 to layer 6, the model performs one more small attention shift (from N[6] to N[8]). We hypothesize the model does that to keep itself attended to the left ring only (instead of both the left and the right rings). This is reasonable because when N[8] receives more redundant information about the right ring, its attention score reduces from 0.48 (the 5-th row) to 0.27 (the 6-th row). The strong focus of the model on a particular substructure is also well demonstrated in Fig. 8. As we can see in the last row, although C[10] (at the last column) contains information about the whole molecular graph, its attention score is still significantly smaller than of the substructure rooted at N[8].

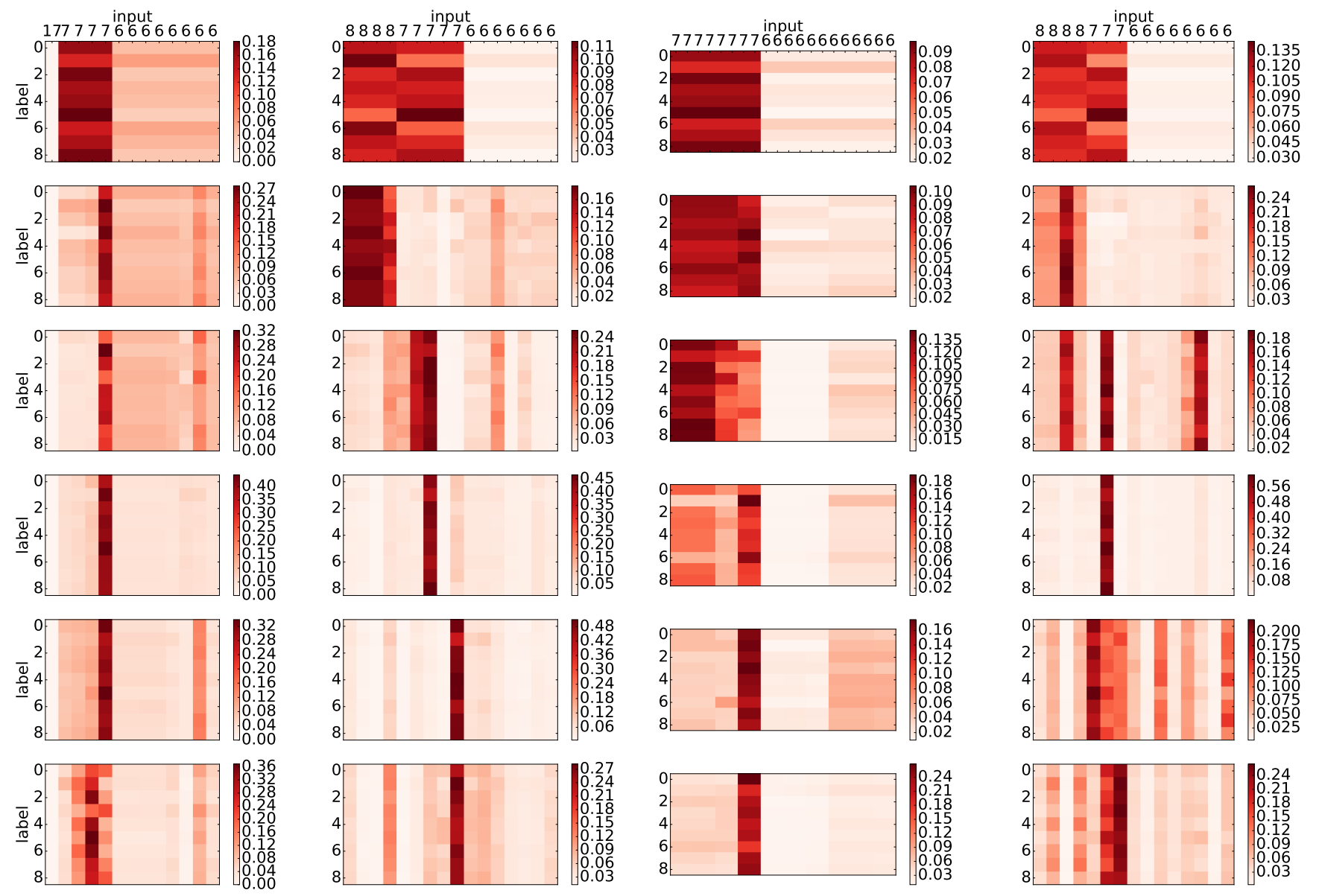

Fig. 7: Normalized label-to-input attention probability at 6 layers of GAML over 4 different molecular graphs sampled from *9cancers*. Darker color refers to higher probability. Columns correspond to input graphs and rows correspond to layers with the first layer drawn on top then the second layer and so on. Each tick in the x-axis is labeled with the atomic number of the corresponding node in the input graph (6: Carbon, 7: Nitrogen, 8: Oxygen, 17: Chlorine). Best view in color.

To discovery typical rooted substructures at a particular depth for a group of classes, we select a node with the best attention score averaged over the present classes for every molecular graph in the training data. Then, we perform clustering on the representation vector of these nodes to find similar substructures. Fig. 9 shows an example of such common substructures shared by different molecules that is typical to all classes in *9cancers*.

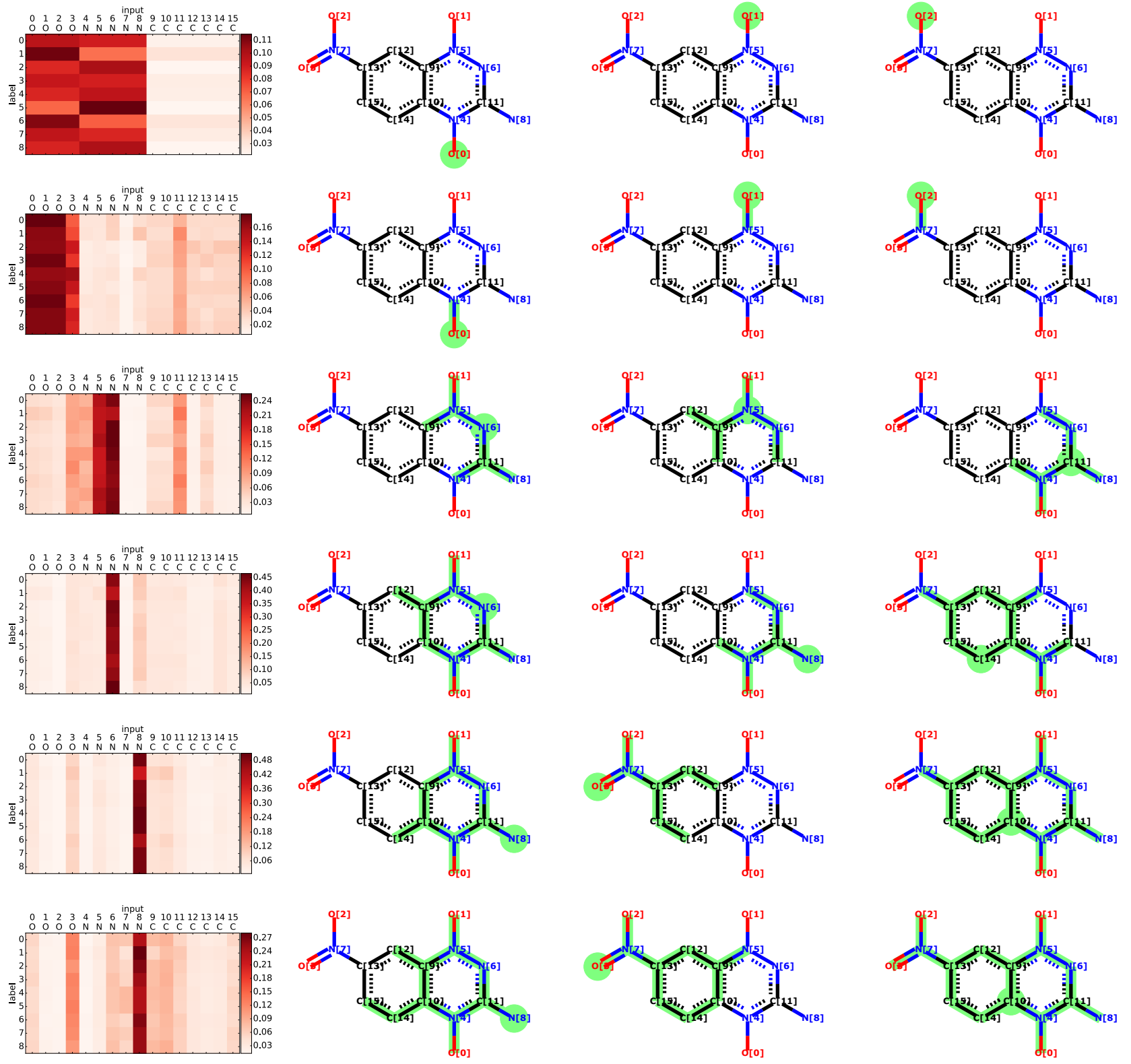

Fig. 8: Attention visualization on substructures of a molecule with PubChem SID of 491286. This molecule is the second example in Fig. 7. Each row specifies the top 3 substructures with the highest attention score (sorted in descending order from left to right) at the corresponding layer. For each substructure at layer $k$, the root atom as well as its neighbor atoms and bonds up to $k$ hops are highlighted in green. Each atom is displayed with its atomic number and its index number (in square brackets) in the molecule. Best view in color.

## 4.2 Multilabel classification with unstructured input

We now test whether our proposed method can work on the traditional setting where the input is a vector.

### *4.2.1 Datasets*

Four datasets are used in this experiments: *media_mill*, *bookmarks*, *Corel5k* and *NUS-WIDE* (see Table 5 for statistics). The former two belong to the text categorization domain where each instance is a document represented as binary bag-of-words. Meanwhile, the latter two belong to the image classification domain where each image is represented as a real-value feature vector. For all datasets, we follow the predefined train/test split so that our results can be comparable to others.

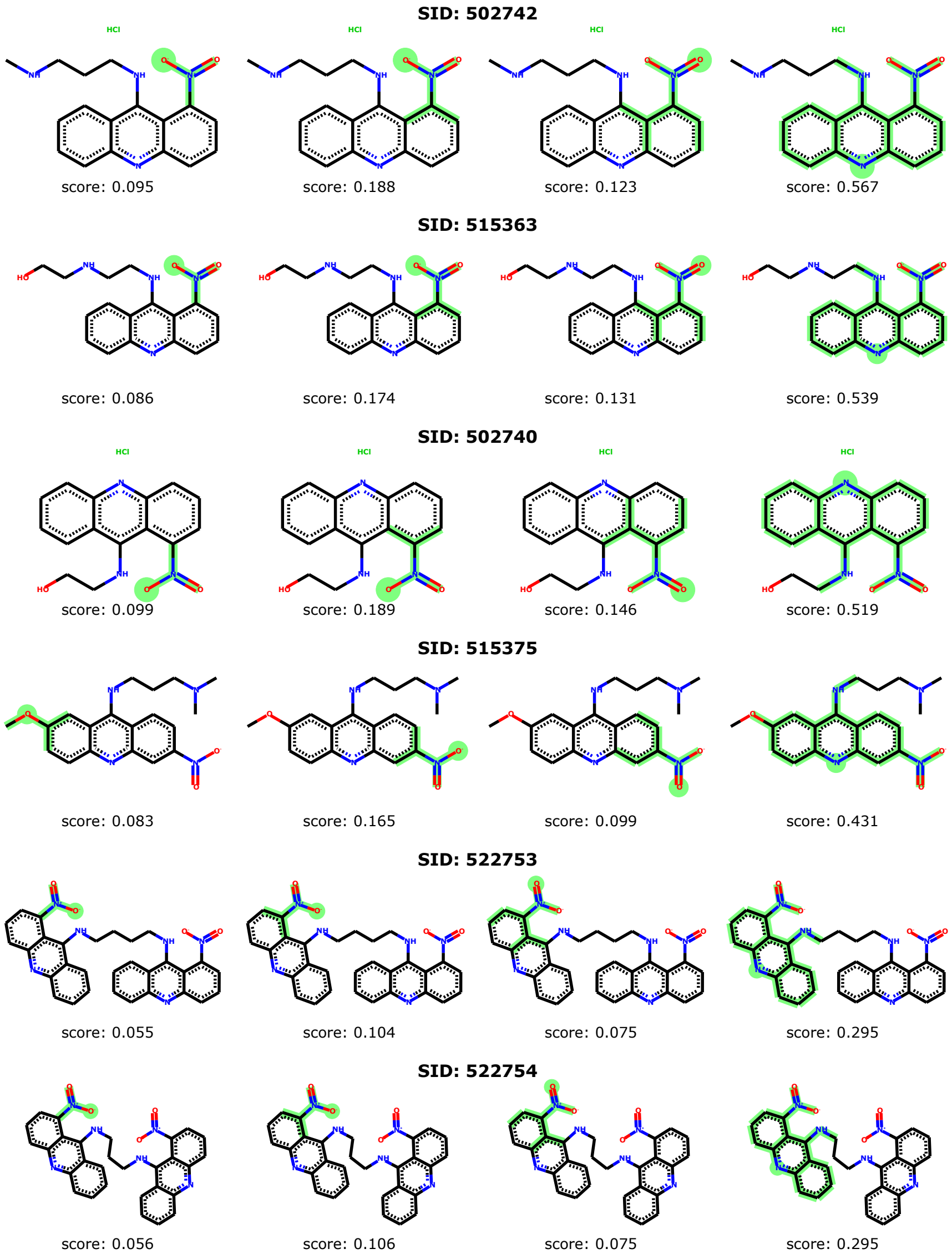

Fig. 9: Common substructures shared by some molecules that are typical to all classes in *9cancers*. Pictures from left to right show the evolution of the rooted substructures with depth from 2 to 5. Along the rows, the molecules are sorted by the average attention scores computed at the topmost layer. Best view in color.

| Dataset | #labels | #features | #total | #train | #test |
|---|---|---|---|---|---|
| *media_mill* | 101 | 120 | 43,907 | 30,993 | 12,914 |
| *bookmarks* | 208 | 2150 | 87,856 | 70,285 | 17,571 |
| *Corel5k* | 374 | 499 | 5,000 | 4,500 | 500 |
| *NUS-WIDE* | 81 | 128 | 269,648 | 161,789 | 107,859 |

Table 5: Statistics of all multilabel datasets with unstructured input.

#### *4.2.2 Baselines*

For comparison, we consider the following methods:

- State-of-the-art classical methods for multilabel classification (MLC) evaluated in [27], which are representative for broader classes of algorithms. They are RAkEL [48] for ensemble methods, ML-kNN [54] for algorithm adaptation methods, HOMER [46] for label power set methods and Calibrated Label Ranking[12] for pairwise ranking. Most of these methods are implemented in well-known multilabel machine learning systems, such as Mulan [47] and Meka [34] with careful hyper-parameters tuning by the authors. Thus, their result are strong and reliable. For presentation compactness, we only report the best results in [27].
- Collective multilabel classification with CRF (CML) [13]. This model can learn pairwise correlations among labels via CRF, hence, should be selected as baseline for comparison. We use the Java implementation of CML released on Github[4] and search for the optimal values of “train.gaussianVariance” in $\{0.01, 0.03, 0.1, 0.3, 1, 3, 10\}$. However, we can only test this model on *media_mill* and *NUS-WIDE* since the other two datasets are not accepted by the implementation.
- A Highway Network (HWN) [40], similar to what described in Section 4.1.

#### *4.2.3 Model setting*

The label embedding size is set to 50 for *NUS-WIDE* and *media_mill*, 75 for *bookmarks* and 30 for *Corel5k*. We project input vector to a low dimensional space by using a single layer neural network with ReLU activation before feeding it to GAML. The size of the projected vector is 55 for *NUS-WIDE*, 75 for *media_mill*, 110 for *bookmarks* and 50 for *Corel5k*. For all datasets, the number of message passing layers is set to 6. In training, the batch size for *NUS-WIDE* is 500 while for the other datasets, it is 100. We use $k$-fold cross validation where $k$ is 9 for *Corel5k* and 5 for other dataset. The optimizer is Adam with an initial learning rate of 0.001. We reduce the learning rate by half if the valid loss does not decrease after 5 consecutive epochs for *Corel5k* and 20 for other datasets. The maximum number of epochs is 300 and the early stopping condition is 4 times of the learning rate decay.

[4] https://github.com/cheng-li/pyramid/wiki/CRF

| Dataset | Metrics | Best in [27] | CML [13] | HWN | GAML |
|---|---|---|---|---|---|
| *media_mill* | m-F1 | 56.3 | 45.78 | 56.73 | **57.53** |
| | M-F1 | 11.3 | 3.95 | 13.50 | **14.17** |
| *bookmarks* | m-F1 | 26.8 | - | 32.51 | **33.33** |
| | M-F1 | 11.9 | - | 20.43 | **21.68** |
| *Corel5k* | m-F1 | **29.3** | - | 15.28 | 22.13 |
| | M-F1 | **4.2** | - | 1.83 | 3.82 |
| *NUS-WIDE* | m-F1 | - | 30.42 | 38.50 | **39.83** |
| | M-F1 | - | 3.84 | 9.31 | **11.38** |

Table 6: The performance in the multi-label classification with unstructured input (m-X: micro average of X, M-X: macro average of X). Missing results are because NUS-WIDE was not tested in [27] and CML implementation did not work on *bookmarks* and *Corel5k*. Bold indicates better values.

#### *4.2.4 Results*

The classification performance of all the methods is presented in Table 6. The deep networks (HWN and GAML) outperform traditional methods and CML on most datasets except for *Corel5k* – the smallest dataset. Especially, on *bookmarks*, our model improves the micro F1 and macro F1 over the best traditional methods by about 7% and 10%, respectively. In the case of *Corel5k*, the best traditional method is CLR (see [27]), a ranking-based method that uses SVM as a base classifier. SVM appears to be more robust than deep networks on small datasets like *Corel5k*. Compared to HWN, GAML achieves better results on all datasets. This supports our model's strength in learning correlations between labels and the input at multiple levels of abstraction.

## 5 Discussion

We introduced GAML, a new graph neural network to tackle an open problem of multi-label learning over graph structured data. The key insight is to realize that label nodes and input nodes can be put into a joint graph to model the multi-way relations among labels and subgraphs. This is achieved through a message passing scheme that exchanges information between connected nodes across multiple steps and an attention mechanism that enables selective flowing of information between label nodes and input nodes. Our model is highly flexible and scalable. We evaluated GAML using an extensive set of experiments on both graph structured and unstructured inputs. Our results clearly demonstrate the efficacy of the proposed model.

This work opens up a wide room for the future at both applied and theoretical fronts. GAML is directly applicable to many other domains. One example is shopping basket recommendation, where users play the role of labels (with or without profile), and item basket modeled as input graph of items. Alternatively, items recommendation to user group works in a similar way, where the user group forms a social graph, and items play the role of labels. At the modeling front, a next step is to extend GAML from label node classification to full graph prediction, where edges are also predicted. Additionally, the current setting is open for auxiliary tasks, e.g., the input graph is node-labeled.